\documentclass[9pt,conference]{IEEEtran}
\IEEEoverridecommandlockouts

\usepackage{amsmath,graphicx}
\usepackage{cite}
\usepackage{colortbl}
\usepackage{comment}
\usepackage{adjustbox}
\usepackage{booktabs}
\usepackage{arydshln}
\usepackage{multirow}

\usepackage{amssymb}
\usepackage{pifont}

\usepackage{graphicx}
\usepackage{amsmath}
\usepackage{amssymb}
\usepackage{booktabs}
\usepackage{algorithm}
\usepackage{listings}
\usepackage{placeins}
\usepackage[accsupp]{axessibility}
\usepackage{nccmath}
\usepackage{authblk}
\makeatletter
\renewcommand\AB@affilsepx{, \protect\Affilfont}
\makeatother
\usepackage{tcolorbox}

\usepackage{caption} 
\usepackage{subcaption}
\usepackage{makecell}

\newcommand{\CC}[1]{\cellcolor{#1}}
\definecolor{ftcolor}{rgb}{1,1,1} 
\definecolor{baselinecolor}{rgb}{1,1,1}
\definecolor{contrcolor}{rgb}{1.0, 0.98, 0.8}
\definecolor{predcolor}{rgb}{0.74, 0.83, 0.9}
\definecolor{decorrcolor}{rgb}{0.98, 0.85, 0.87} 
\definecolor{knowcolor}{rgb}{0.80, 0.94, 0.75}
\definecolor{offlinecolor}{rgb}{1,1,1}
\definecolor{firsttaskcolor}{rgb}{0.97, 0.97, 0.97}
\definecolor{azure(colorwheel)}{rgb}{0.0, 0.5, 1.0}
\definecolor{gray(x11gray)}{rgb}{0.75, 0.75, 0.75}
\definecolor{lightgray}{rgb}{0.90, 0.90, 0.90}
\definecolor{darkgray}{rgb}{0.66, 0.66, 0.66}
\definecolor{teagreen}{rgb}{0.82, 0.94, 0.75}
\definecolor{almondlow}{RGB}{252,239,219} 
\definecolor{almondmiddle}{RGB}{237,225,206} 
\definecolor{almondhigh}{RGB}{224,213,194} 
\definecolor{almondultra}{RGB}{214,204,186} 
\definecolor{greylow}{RGB}{235,235,235} 
\definecolor{greymiddle}{RGB}{211,211,211} 
\definecolor{greyhigh}{RGB}{192,192,192} 
\definecolor{pinksecondbest}{RGB}{252, 241, 241}

\definecolor{pastelred}{RGB}{232, 131, 131}
\definecolor{pastelviolet}{rgb}{0.8, 0.7, 0.79}
\definecolor{champagne}{rgb}{0.97, 0.91, 0.81}

\NewDocumentCommand\emojice{}{
    \includegraphics[scale=0.009]{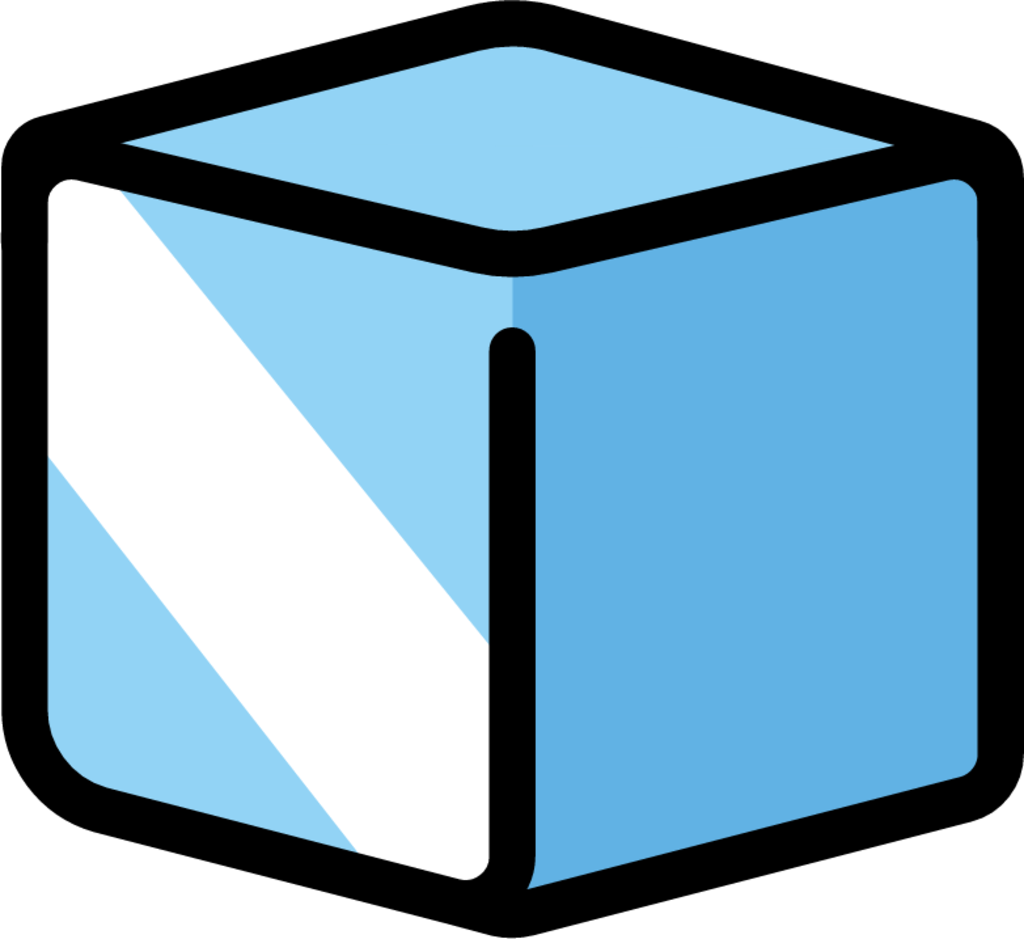}
}

\NewDocumentCommand\emojifire{}{
    \includegraphics[scale=0.009]{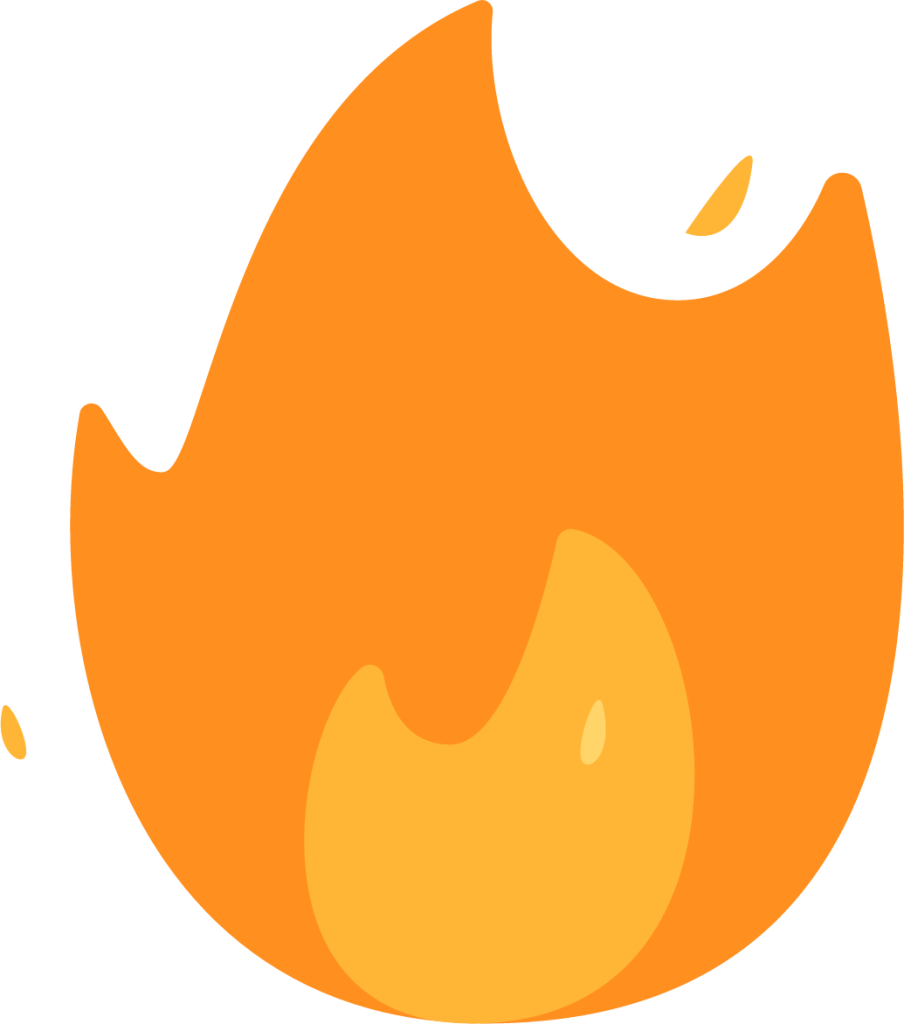}
}

\DeclareSymbolFont{extraup}{U}{zavm}{m}{n}
\DeclareMathSymbol{\varheart}{\mathalpha}{extraup}{86}
\DeclareMathSymbol{\vardiamond}{\mathalpha}{extraup}{87}
\usepackage[T1]{fontenc}

\usepackage[utf8]{inputenc}

\usepackage{microtype}

\usepackage{inconsolata}

\title{Large Language Models are Strong Audio-Visual Speech Recognition Learners}
\author[$\heartsuit$,$\vardiamond$]{\textbf{Umberto Cappellazzo}$^{\textbf{\dag},}$\thanks{$^\textbf{\dag}$Work done while visiting Imperial College London. $\clubsuit$ Funded by the European Union’s Horizon 2020 project ELOQUENCE (grant 101070558).\\Only non-Meta authors conducted any of the dataset preprocessing (no dataset pre-processing took place on Meta’s servers or facilities).}}
\author[$\spadesuit$]{\textbf{Minsu Kim}}
\author[$\spadesuit$]{\textbf{Honglie Chen}}
\author[$\spadesuit$]{\textbf{Pingchuan Ma}}
\author[$\spadesuit$,$\heartsuit$]{\textbf{Stavros Petridis}}
\author[$\clubsuit$]{\textbf{Daniele Falavigna}}
\author[$\clubsuit$]{\authorcr \textbf{Alessio Brutti}}
\author[$\spadesuit$,$\heartsuit$]{\textbf{Maja Pantic}}

\affil[$\heartsuit$]{Imperial College London}
\affil[$\vardiamond$]{University of Trento}
\affil[$\spadesuit$]{Meta AI}
\affil[$\clubsuit$]{Fondazione Bruno Kessler}

\begin{document}
%
\maketitle
\begin{abstract}
Multimodal large language models (MLLMs) have recently become a focal point of research due to their formidable multimodal understanding capabilities. For example, in the audio and speech domains, an LLM can be equipped with (automatic) speech recognition (ASR) abilities by just concatenating the audio tokens, computed with an audio encoder, and the text tokens to achieve state-of-the-art results. On the contrary, tasks like visual and audio-visual speech recognition (VSR/AVSR), which also exploit noise-invariant lip movement information, have received little or no attention. To bridge this gap, we propose \texttt{Llama-AVSR}, a new MLLM with strong audio-visual speech recognition capabilities. It leverages pre-trained audio and video encoders to produce modality-specific tokens which, together with the text tokens, are processed by a pre-trained LLM (e.g., Llama3.1-8B) to yield the resulting response in an auto-regressive fashion. \texttt{Llama-AVSR} requires a small number of trainable parameters as only modality-specific projectors and LoRA modules are trained whereas the multi-modal encoders and LLM are kept frozen. We evaluate our proposed approach on LRS3, the largest public AVSR benchmark, and we achieve new state-of-the-art results for the tasks of ASR and AVSR with a WER of 0.79\% and 0.77\%, respectively. To bolster our results, we investigate the key factors that underpin the effectiveness of \texttt{Llama-AVSR}: the choice of the pre-trained encoders and LLM, the efficient integration of LoRA modules, and the optimal performance-efficiency trade-off obtained via modality-aware compression rates. 
\end{abstract}
\section{Introduction}
\label{sec:intro}

By integrating both auditory and visual data, audio-visual speech recognition (AVSR) aims to enhance the capabilities of speech recognition systems \cite{afouras2018deep, av-hubert, raven, autoavsr}. Notably, the additional use of visual lip movement information is highly beneficial in environments characterized by background noise or ambient speech, enhancing noise robustness \cite{shi2022robust, gong2023whisper}. Multiple works have shown that having access to a wide amount of labeled audio-visual data is the key to obtaining strong results \cite{ViT3D-CM, lpconformer}. However, such a reliance on large-scale transcribed datasets of up to $100$K samples is prohibitively expensive and time-consuming. 
Therefore, recent methods have focused on different paradigms. A vast class of works relies on the Self-Supervised Learning (SSL) paradigm \cite{gui2024survey} by pre-training on large-scale datasets of unlabeled videos and then fine-tuning on a few hundred hours of labeled videos \cite{av-hubert, raven, u-hubert, Av-data2vec, braven}. Another line of research \cite{vsr-wild, autoavsr, yeo2024visual} proposes to use publicly-available pre-trained ASR models to automatically annotate large-scale audio-visual datasets. 


Large language models (LLMs) have shown exceptional abilities in handling natural language tasks \cite{mistral, llama3}. Their impressive generalization and instruction-following capabilities have spurred the development of multimodal LLMs (MLLMs) \cite{mm1, cambrian}, thus making it possible to process other modalities rather than just text. MLLMs have obtained remarkable performance in several tasks like vision-language \cite{flamingo, blip2, liu2024visual, improved}, video understanding \cite{videollama2, llama-vid}, and audio understanding \cite{pengi, audio-flamingo, gama} to name a few. Some works also propose unified MLLMs that handle multiple modalities at the same time \cite{macaw, uni-moe, crema}. 

Recent research has also highlighted the effectiveness of LLMs for the task of automatic speech recognition (ASR) \cite{chen2024s, hu2024large, ma2024embarrassingly, yu2024connecting, fathullah2024prompting} and visual speech recognition \cite{VSP-LLM}. An LLM can be endowed with speech recognition abilities by conditioning on the audio embeddings and by training lightweight projector layers \cite{ma2024embarrassingly} or LoRA modules \cite{fathullah2024prompting}. Given the excellent performance achieved by LLM-empowered ASR models and the ability of LLMs to process multiple modalities simultaneously \cite{uni-moe, crema}, we aim to explore whether LLMs can be adapted to perform the task of AVSR. In this way, the LLM would rely on multi-modal tokens that convey the same information but from complementary perspectives. For example, the LLM can highly benefit from additional video tokens in the presence of noisy acoustic environments. Furthermore, there is a notable absence of comprehensive research that systematically explores the integration of ASR, VSR, and AVSR tasks using LLMs. Driven by these considerations, we pose the following research question:

\begin{tcolorbox}[colback=green!5]
\textbf{(}$\mathbf{Q}$\textbf{)} \textit{How can we leverage powerful \textbf{LLMs} to carry out the tasks of audio, visual, and audio-visual speech recognition?}
\end{tcolorbox}

To tackle \textbf{(}$\mathbf{Q}$\textbf{)}, we propose a new framework, \texttt{Llama-AVSR}, which harnesses 
pre-trained LLMs (e.g., Llama-based \cite{llama2, llama3}) and audio/video encoders for the tasks of ASR, VSR, and AVSR. Our approach involves extracting modality-specific feature representations from pre-trained encoders. These features are then downsampled 
to reduce computational complexity and projected into the LLM's embedding space using lightweight projectors, resulting in audio/video tokens. By concatenating these tokens with the text tokens, we 
integrate information from all modalities. These combined tokens are 
fed into the LLM that generates the transcriptions in an auto-regressive way. 
\texttt{Llama-AVSR} only trains the projectors and the LLM's LoRA module while keeping the pre-trained encoders and LLM frozen. In this way, we significantly reduce the number of trainable parameters compared to traditional methods that train the entire pipeline. Moreover, the modularity of our \texttt{Llama-AVSR} framework facilitates the seamless integration of various pre-trained encoders and LLMs of different sizes. This flexibility allows us to easily adapt our model to meet specific requirements of the size-performance trade-off.

\texttt{Llama-AVSR} achieves new state-of-the-art results on the LRS3 dataset for the tasks of ASR ($\textbf{0.79}$\%) and AVSR ($\textbf{0.77}$\%) by training only $42$ and $57$ million parameters, respectively. For VSR, our method outperforms prior works using LLM \cite{VSP-LLM} and attains comparable performance  with state-of-the-art methods \cite{raven, autoavsr}. Moreover, we experimentally analyze the \textbf{key factors} that lead to the effectiveness of \texttt{Llama-AVSR} for the three tasks, observing that: \textbf{\ding{182})} the choice of the pre-trained audio and video encoders as well as of the LLM has a big impact on the final performance, \textbf{\ding{183})} incorporating LoRA modules into the LLM and video encoder are highly beneficial to improve the overall performance whilst requiring limited additional parameters, 
\textbf{\ding{184})} the selection of the compression rate is crucial to finding the optimal trade-off between performance and efficiency for all three tasks.
\section{Method}
\label{sec:method}
\begin{figure}[t]
    \centering
    \includegraphics[width=7cm]{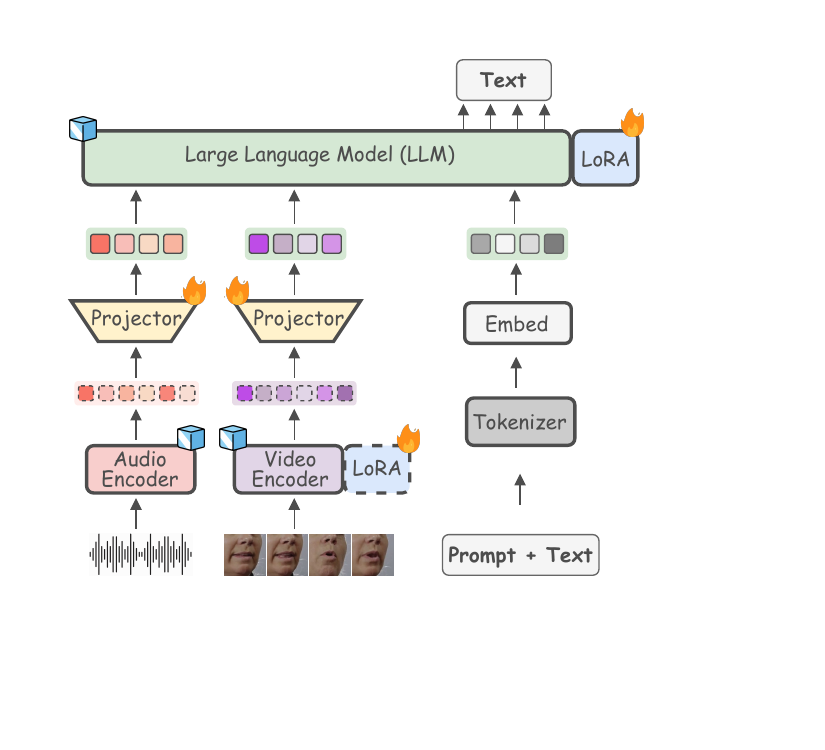}
    \caption{Illustration of \texttt{Llama-AVSR}. Audio and video features are extracted via pre-trained encoders and subsequently downsampled and projected into the LLM space through modality-specific projectors. The resulting audio and video tokens are concatenated with the textual ones and processed by the LLM. The video encoder is equipped with the LoRA module only for the VSR task (dashed outline). \emojifire and \emojice means that the block is trained and kept frozen, respectively.}
    \label{fig:pipeline}
    \vspace{-0.6cm}
\end{figure}

Our method, \texttt{Llama-AVSR}, leverages the capabilities of pre-trained audio and video encoders as well as LLMs for carrying out the tasks of ASR, VSR, and AVSR. It comprises three main components: \textbf{1)} \textit{modality-specific pre-trained encoders} (i.e., audio and video), \textbf{2)} \textit{modality-specific projectors}, and \textbf{3)} an \textit{LLM}. This architecture is referred to as \textbf{Multimodal LLM} (MLLM) as the LLM produces text responses in an auto-regressive way given a sequence of multimodal input tokens (audio/video + text). Our approach is illustrated in Figure \ref{fig:pipeline}. Due to its versatility and streamlining pipeline, we adopt the \textit{decoder-only}-based approach \cite{liu2024visual, vila, mm1}, which combines pre-trained LLMs and multimodal inputs through lightweight connectors, rather than the \textit{cross-attention}-based approach \cite{flamingo, cambrian, llama3}. We delve into the details of each MLLM's building block below.


\textbf{Modality-specific Pre-trained Encoders}. We exploit pre-trained audio (e.g., Whisper \cite{whisper}) and video (e.g., AV-HuBERT \cite{av-hubert}) encoders to extract meaningful audio and video features to be harnessed by the LLM. These pre-trained encoders are maintained frozen throughout the training process. For the task of VSR only, we found empirically that adding a LoRA module to the video encoder brings additional improvements at a small overhead cost of around $6$M parameters.

\textbf{Modality-specific Projector}. The projector, sometimes called connector \cite{llama-vid, uni-moe}, bridges the pre-trained encoders and the LLM by translating audio and visual features into understandable tokens that the LLM can process. The quality of these tokens significantly influences the MLLM's performance. Furthermore, the projector plays a crucial role in terms of efficiency since it determines how many tokens will be processed by the LLM, which handles most of the computational load. For instance, $6$ seconds of audio-visual features yield $450$ frames, thus processing these long sequences with the LLM presents substantial computational challenges. Given their simplicity and popularity, we opt to employ linear projectors, which excel at capturing fine-grained details by preserving local audio and visual patterns without loss due to their inherent frame-wise transformation \cite{honeybee}. However, they usually struggle in terms of efficiency and scalability as the number of features (before the projector) and tokens (after) remains constant. We tackle this by downsampling the audio and video features to reduce their sequence length: we first concatenate $K$ (we call it \textit{compression rate}) consecutive features along the hidden dimension (i.e., the sequence length/hidden size is reduced/increased by a factor $K$), and then the projector maps the audio/video features into audio/video tokens through two linear layers to match the text tokens hidden size. Finally, audio and video tokens are concatenated with the textual tokens.

\textbf{LLM}. The goal of the LLM is to generate instruction-following responses given a sequence of multimodal inputs. In addition to the text tokens, the LLM processes: 1) audio tokens for the ASR task, 2) video tokens for the task of VSR, and 3) both audio and video tokens for the task of AVSR. Therefore, the LLM digests audio and/or video and text (instruction/prompt + transcription) tokens while generating the text response (i.e., the transcription) in an auto-regressive fashion. Formally, if we consider the task of AVSR, the multimodal input comprises audio $\mathbf{X}_{\mathsf{aud}}$, video $\mathbf{X}_{\mathsf{vid}}$, and text $\mathbf{X}_{\mathsf{text}}$ tokens. The LLM predicts the response $\mathbf{Y} = \{y_i\}_{i=1}^{N}$ conditioned on the multimodal input tokens, where $N$ represents the number of tokens. Accordingly, the probability of the target response $\mathbf{Y}$ is computed by:
\begin{equation}
p(\mathbf{Y}|\mathbf{X}_{\mathsf{aud}}, \mathbf{X}_{\mathsf{vid}}, \mathbf{X}_{\mathsf{text}}) = \prod_{i=1}^{N}p(y_i|\mathbf{X}_{\mathsf{aud}}, \mathbf{X}_{\mathsf{vid}}, \mathbf{X}_{\mathsf{text}}, y_{<i}),
\end{equation}
where $y_{<i}$ is the generated output sequence up to token $i-1$. In all our experiments the LLM is kept frozen while we add a LoRA module to align the LLM responses with the multimodal inputs. 

\section{Experiments and Discussion}
\label{sec:experiments}

\subsection{Implementation Details}

\textbf{Datasets}. We conduct our experiments on LRS3 \cite{LRS3}, the largest publicly available dataset for audio-visual speech recognition. It includes $433$ hours of transcribed English video clips from TED talks. We also report the results when training on only the 30-hour ``trainval'' set of LRS3 (denoted ``low-resource'' setting in \cite{raven}). Additionally, following \cite{autoavsr}, we use the pre-trained Whisper model \cite{whisper} to generate the transcriptions of the English-speaking videos from VoxCeleb2 \cite{VoxCeleb2}, resulting in additional $1,326$ hours. We also report the results from this setting: LRS3 + VoxCeleb2 ($1,756$ hours). 

\textbf{Pre-processing}. We follow \cite{autoavsr} for the pre-processing of the datasets. For the video part, we crop the mouth region of interests (ROIs) through a bounding box of 96 × 96. Each frame is normalised by subtracting the mean and dividing by the standard deviation of the training set. For audio, we only apply z-normalisation per utterance.

\textbf{Training/Inference Details}.  We augment visual inputs through horizontal flipping, random cropping, and adaptive time masking, whereas for audio we only apply adaptive time masking. For training, similar to \cite{autoavsr}, we sample bubble noise from the NOISEX dataset \cite{noisex} using a uniform distribution from the range [-$5$, $0$, $5$, $10$, $15$, $20$, $\infty$]dB and add it to the clean speech signal. For our experiments, we use Whisper-medium \cite{whisper} and 
AV-HuBERT Large \cite{av-hubert} as pre-trained audio and video encoders. These encoders are kept frozen, and only for the VSR task do we add a LoRA module to the video encoder (we use a rank = $64$, resulting in roughly $6$M additional parameters). For AV-HuBERT, we use the checkpoint pre-trained on LRS3 + VoxCeleb2. The projectors consist of two linear layers with ReLU activation in between ($\sim 15$M parameters). For the tasks of ASR and VSR, we apply a compression rate $K$ of $3$ for the settings with $433$ and $1756$ hours, and $2$ for the low-resource setting with $30$ hours. For the task of AVSR, we set $K$ equal to $4$ and $2$ for the audio and video tokens, respectively. We refer the reader to Section \ref{sec:ablations} for a detailed analysis of $K$. For the main experiments, we use the pre-trained Llama3.1-8B \cite{llama3} as our LLM, while for the ablation studies we also experiment with TinyLlama \cite{tinyllama}, Llama2-7B \cite{llama2}, and Llama2-13B \cite{llama2} to compare LLMs of different sizes. The LLM is coupled with a LoRA module ($\sim 27$M parameters). The prompt for the LLM is ``\texttt{Transcribe \{\textbf{task\_prompt}\} to text.}'', where \texttt{\textbf{task\_prompt}} $\in$ \{``\texttt{speech}'', ``\texttt{video}'', ``\texttt{speech and video}''\} depending on the task we study. We train our model for $10$ epochs with the AdamW optimizer with cosine annealing scheduler and weight decay set to $0.1$. The learning rate is set to 1e-3 for ASR and AVSR tasks, and 5e-4 for VSR. For decoding, we use beam search with a beam width of $15$ and temperature of $0.6$. 

\subsection{Main Results}
\begin{table}
\renewcommand{\arraystretch}{1.2}
\renewcommand{\tabcolsep}{1.1mm}
\centering
\caption{WER (\%) of \texttt{Llama-AVSR} and prior works on the LRS3 dataset. We report results based on the task (ASR, VSR, AVSR) and on the  labeled hours ($30$, $433$, $1756$).} 
\resizebox{0.999\linewidth}{!}{
\begin{tabular}{llccc}
    \toprule
     \textbf{Method} & \textbf{Encoder(s)} & \makecell{\textbf{Trainable}\\ \textbf{Par. (M)}} & \makecell{\textbf{Labeled}\\ \textbf{Hours}}  & \textbf{WER} $\downarrow$ \\
    \midrule
     \multicolumn{5}{c}{\CC{predcolor} \textbf{\textit{Audio-Only Setting}}} \\
     RAVEn \cite{raven} &Transformer & 328 & 30/433 & 1.9/1.4 \\
     BRAVEn \cite{braven} & Transformer & 328 & 30/433 & 1.7/1.1 \\
     CM-seq2seq \cite{cm-seq2seq} &Conformer & 250 &433 & 2.3 \\
     Fast Conformer \cite{burchi2024multilingual} & Conformer & 197 & 435 & 1.6 \\
     AV-HuBERT \cite{av-hubert} & Transformer & 325 & 433 & 1.3 \\
     Whisper-finetuned \cite{whisper-flamingo} &Whisper & 1550 & 433 & 2.3 \\
     auto-avsr \cite{autoavsr} &Conformer & 243 & 1902/3448 & 1.0/1.0 \\
     \hdashline \addlinespace[2pt]
     \textbf{\texttt{Llama-AVSR}} & AV-HuBERT A& 40 & 433 & 1.4 \\
     \textbf{\texttt{Llama-AVSR}} & Whisper & \textbf{42} &30 & \textbf{1.5} \\
     \textbf{\texttt{Llama-AVSR}} & Whisper & \textbf{42} &433 & \textbf{1.1} \\
    \rowcolor{teagreen}
    \textbf{\texttt{Llama-AVSR}} & Whisper & \textbf{42} & 1756 & \textbf{0.79} \\ \hline
     
    \multicolumn{5}{c}{\CC{champagne} \textbf{\textit{Video-Only Setting}}} \\
    RAVEn \cite{raven} & Transformer & 328 & 30/433 & 24.8/24.4  \\
    BRAVEn \cite{braven} & Transformer & 328 & 30/433 & 20.0/20.1 \\
    AV-data2vec \cite{Av-data2vec} & Transformer & 325 & 30/433 & 30.8/28.5 \\
    auto-avsr \cite{autoavsr} &Conformer & 250 & 433 & 36.3 \\
    AV-HuBERT \cite{av-hubert} & Transformer & 325 & 433 & 26.9 \\
    VSP-LLM \cite{VSP-LLM} &AV-HuBERT V & 17 & 433 &26.7 \\
    auto-avsr \cite{autoavsr} & Conformer &250 & 1902 & 23.5 \\
    LP Conformer \cite{lpconformer} & Conformer & 570 & \textbf{\textcolor{red}{100K}} &12.8 \\ 
    \hdashline \addlinespace[2pt]
    \textbf{\texttt{Llama-AVSR}} & AV-HuBERT V& \textbf{48} & 30 & \textbf{28.4} \\
    \textbf{\texttt{Llama-AVSR}} & AV-HuBERT V& \textbf{48} & 433 & \textbf{25.3} \\
    \rowcolor{teagreen}
    \textbf{\texttt{Llama-AVSR}} & AV-HuBERT V& \textbf{48} &1756 & \textbf{24.0} \\
    \hline
    \multicolumn{5}{c}{\CC{pastelviolet} \textbf{\textit{Audio-Visual Setting}}}\\
    CM-seq2seq \cite{cm-seq2seq} & Conformer & 250 &433 & 2.3 \\
    Whisper-Flamingo \cite{whisper-flamingo} & Whisper & 631 & 433 & 1.5 \\
    CMA \cite{CMA} &Transformer &500 & 433 & 1.5 \\
    auto-avsr \cite{autoavsr} & Conformer & 425 &1902/3448 & 1.0/0.9 \\
    Fast Conformer \cite{burchi2024multilingual} & Conformer & 197 & 1687 & 0.9 \\
    ViT3D-CM \cite{ViT3D-CM} & Transformer & / & \textbf{\textcolor{red}{90K}} & 1.6 \\
    LP Conformer \cite{lpconformer} & Conformer & 570 & \textbf{\textcolor{red}{100K}} &0.9 \\ 
    \hdashline \addlinespace[2pt]
    \textbf{\texttt{Llama-AVSR}} & AV-HuBERT AV & 59 & 433 & 1.3 \\
    \textbf{\texttt{Llama-AVSR}} & Whisper + AV-HuBERT V & \textbf{57} & 433 & \textbf{0.95} \\
    \rowcolor{teagreen}
    \textbf{\texttt{Llama-AVSR}} & Whisper + AV-HuBERT V& \textbf{57} &1756 & \textbf{0.77} \\
     
 \bottomrule
\end{tabular}}
\label{tab:main}
\vspace{-0.5cm}
\end{table}

We report the results in terms of Word Error Rate (WER) obtained by \texttt{Llama-AVSR} in Table \ref{tab:main}. We compare our approach with multiple state-of-the-art works based on the considered task (ASR, VSR, AVSR) and the number of labeled hours. We also include the number of trainable parameters and the encoder(s) employed by each method, which may utilize a pre-trained model (as in our case, VSP-LLM \cite{VSP-LLM} and Whisper-Flamingo \cite{whisper-flamingo}) or be trained from scratch \cite{autoavsr, av-hubert, lpconformer} (i.e., Transformer or Conformer). For the task of ASR (``audio-only setting''), \texttt{Llama-AVSR} sets a new state-of-the-art with a WER of $0.79$\% by training on LRS3 + VoxCeleb2 ($1756$ hours), and it also outperforms the other methods when using $433$ and $30$ hours achieving WER results of $1.5$\% and $1.1$\%, respectively. Remarkably, our approach only requires $\textbf{42}$M trainable parameters, which are far fewer compared to all the other methods. We also report the WER achieved by fine-tuning completely a Whisper encoder-decoder model on LRS3 as reported in \cite{whisper-flamingo}. Not only does this approach attain a WER much higher than \texttt{Llama-AVSR} ($2.3$\% vs $1.1$\%), but it also requires the updates of more than $1.5$B parameters. Moreover, their model harnesses Whisper-Large while \texttt{Llama-AVSR} uses the Medium-size version. We also report the case in which AV-HuBERT is used as the 
audio encoder in place of Whisper (``AV-HuBERT A'').

For the video-only setting, \texttt{Llama-AVSR} outstrips VSP-LLM, which is the only prior LLM-based method that exploits a pre-trained video encoder (AV-HuBERT) and LLM (Llama2-7B) and adds an extra task during training (visual speech translation). In addition to this, when we train on $433$ hours we see that \texttt{Llama-AVSR} outperforms various methods like auto-avsr and AV-HuBERT. However, the gap with methods like RAVEN and BRAVEn is more noticeable. We speculate that more sophisticated projectors such as those proposed in recent works \cite{videollama2, honeybee} could produce more fine-grained and expressive tokens and so bridge this gap, yet we leave it for future works. We also observe that adding more training data results in further improvement, achieving performance parity with respect to auto-avsr and RAVEn. 

Finally, for the task of AVSR, we obtain two new SOTA of $0.95$\% and $0.77$\% when using $433$ and $1756$ hours. This is achieved by using Whisper as the audio encoder and AV-HuBERT as the video encoder. If, instead, we use AV-HuBERT to process both audio and video (``AV-HuBERT AV''), the results are slightly worse, in line with what we observed for the ASR task. We point out that our model outperforms methods that exploit tens of thousands of hours \cite{ViT3D-CM, lpconformer}. By comparing with the ASR results, we notice that the additional use of video data is more helpful when we train on LRS3 only ($1.1\% \to 0.95$\%), whereas the gain is reduced when more data are available ($0.79\% \to 0.77$\%), and this trend in line with previous works \cite{Av-data2vec, autoavsr}. 

\subsection{Analysis on the Key Factors of \texttt{Llama-AVSR}}
\label{sec:ablations}
\begin{figure}[t]
\centering
   
\includegraphics[width=8cm]{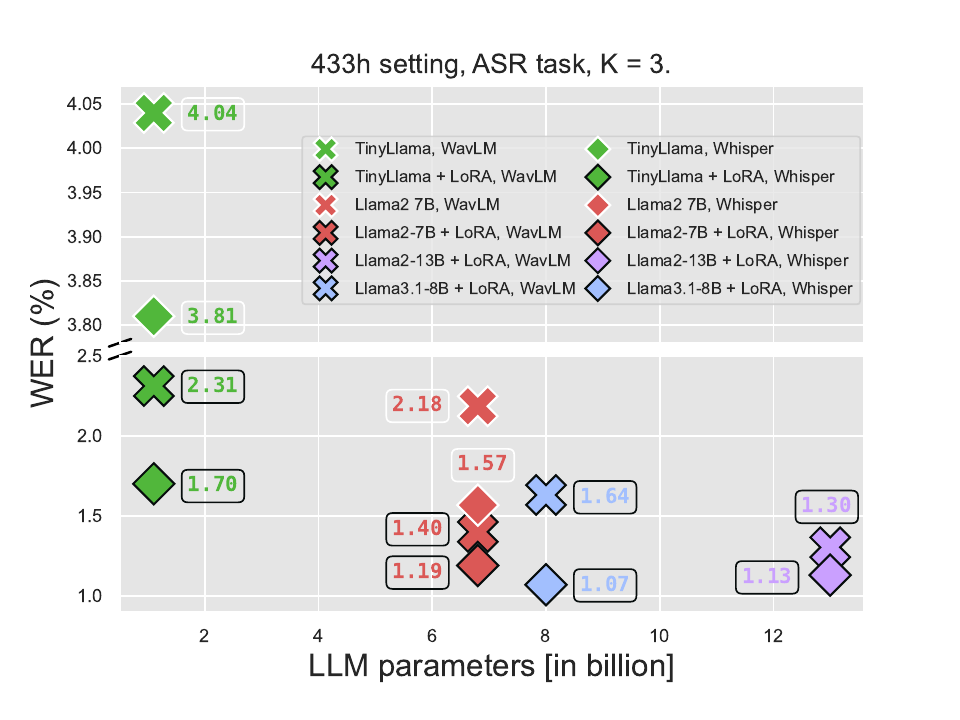}

\caption{WERs of multiple \texttt{Llama-AVSR} configurations for ASR task.}
\label{fig:LLMs_WER}
\vspace{-0.5cm}
\end{figure}

\begin{table}
\renewcommand{\arraystretch}{1.2}
\renewcommand{\tabcolsep}{3mm}
\centering
\caption{WER results for different configurations of \texttt{Llama-AVSR} for the VSR task. ``/'' means LoRA is not used.} 
\resizebox{0.9\linewidth}{!}{
\begin{tabular}{lccc}
\toprule
 \multirow{2}{*}{\textbf{Encoder}} & \multirow{2}{*}{\textbf{LoRA Position}} & \multicolumn{2}{c}{\textbf{LLM}} \\
\cmidrule(l){3-4} & & \cellcolor{pastelviolet}\textbf{Tiny-Llama}&\cellcolor{predcolor} \textbf{Llama3.1-8B} \\
\midrule

AV-HuBERT & / & 30.2 & 28.4 \\
AV-HuBERT & LLM & 29.2 & 26.9 \\
\rowcolor{teagreen}
AV-HuBERT & LLM + enc & \textbf{28.3} & \textbf{25.3} \\
\hdashline \addlinespace[2pt]
RAVEn & LLM + enc & \textcolor{red}{\textbf{34.2}} & \textcolor{red}{\textbf{32.4}}\\
 \bottomrule
 \label{tab:vsr_ablation}
\end{tabular}}
\vspace{-0.7cm}
\end{table}

\textbf{Exploring Different Encoders and LLMs for \texttt{Llama-AVSR}}. We conduct multiple ablation studies on various components of \texttt{Llama-AVSR} to understand their impact on the final performance. We focus on the setting with $433$ hours. For the ASR task, we study four different LLMs from the Llama family with increasing size: \textit{TinyLlama} ($1.1$B parameters), \textit{Llama2-7B}, \textit{Llama2-13B}, and the more recent \textit{Llama3.1-8B}. These LLMs are depicted in different colors in Figure \ref{fig:LLMs_WER}. In addition to this, we ablate the choice of the pre-trained audio encoder, and we compare \textit{Whisper-medium} \cite{whisper} with \textit{WavLM Large} \cite{wavlm} (diamond and cross markers, respectively). Both encoders have around $300$M parameters. We also report the WER obtained \textit{with} (black outline's marker) and \textit{without} (white outline's marker) LoRA at the LLM side. In Figure \ref{fig:LLMs_WER}, we can observe the following trends: \textbf{1)} adding LoRA to the LLM is highly beneficial regardless of the LLM and encoder used. For example, for TinyLlama + WavLM/Whisper, we can improve the performances from WERs of $4.04$/$3.81$\% to $2.31$/$1.70$\%. A similar trend is noticed for Llama2-7B, albeit the gain is reduced. \textbf{2)} The choice of the encoder is also crucial, and this trend becomes more evident when LoRA is not used. This is because better encoded representations enhance the LLM's ability to comprehend, especially in scenarios without LoRA, which typically assists the LLM in aligning the encoded speech features with the pre-learned textual space. \textbf{3)} We observe similar performance when LoRA and Whisper are used among the three biggest LLMs, although LLama3.1-8B surpasses Llama2-13B while using almost half as many parameters. 
Finally, we highlight how a smaller and less powerful LLM such as TinyLlama can surpass Llama2-7B and approach Llama3.1-8B when the right configuration is set. 

For VSR, we compare two video encoders: \textit{AV-HuBERT} \cite{av-hubert} and \textit{RAVEn} \cite{raven}. Since the LoRA module is applied both to the LLM and video encoder, we study $3$ configurations: \textbf{1)} no LoRA module is applied, \textbf{2)} the LoRA module is added to the LLM, and \textbf{3)} LoRA modules are applied to both. Table \ref{tab:vsr_ablation} shows that applying LoRA modules to both leads to gains of almost $3$ points for both TinyLlama and Llama3.1-8B. 
Furthermore, AV-HuBERT achieves much better performance than RAVEn. This could be attributed to the fact that the hidden units discovered through clustering SSL features (e.g., HuBERT \cite{hsu2021hubert}, AV-HuBERT \cite{av-hubert}) are closely related to linguistic information, such as phonemes. This relationship may allow the LLM to more easily adapt the representations of cluster-based SSL models. 

\begin{figure}[t]
    \centering
    \includegraphics[width=6cm]{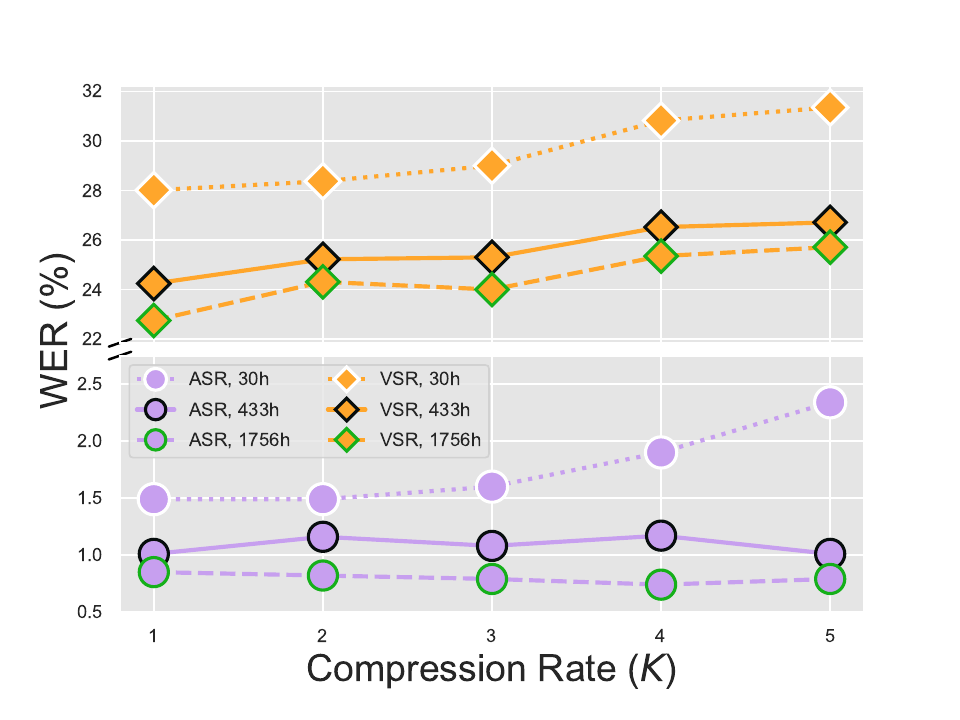}
    \caption{WER trend as a function of $K$ for the ASR and VSR tasks.}
    \label{fig:downsampling}
\end{figure}

\textbf{Efficiency-Performance Trade-off for ASR and VSR Tasks}. Since most of the computational load lies in the LLM, it is quite common to reduce the number of tokens. At the same time, the loss in resolution 
inevitably results in a performance's drop. Consequently, it is crucial to find a compromise in terms of efficiency and performance. In this direction, we report the WER by varying the compression rate $K$ in the range of [1-5] for both the ASR and VSR tasks. For ASR, a compression rate of $1$ (i.e., no compression) means the LLM processes on average $584$ audio tokens, while the LLM processes $117$ tokens for $5$. 
Figure \ref{fig:downsampling} shows that for the ASR task we can compress the audio tokens up to a factor 5 without impacting performance, consistently reducing the number of tokens processed by the LLM. Reasonably, for the 30h setting the pooling factor is more crucial as the model is trained on substantially less data. For VSR, however, increasing $K$ leads to gradually worse results across all settings (the WER increase is around $2.5$-$3$ points when pushing $K$ to $5$), indicating the need for a more careful balance between efficiency and performance.

\textbf{AVSR Ablation Studies}. We study the optimal audio and video compression rates for the AVSR task ($433$h setting). We use Whisper and AV-HuBERT as audio and video encoders. We point out that the temporal resolution of Whisper's output features is $50$ fps, whereas that of video features is $25$ fps. 
We report the results when tested with different acoustic noise levels as well as in a clean setting (SNR level = $\infty$) in Table \ref{tab:ablation_AVSR}. Similar to \cite{autoavsr}, we inject babble noise from the NOISEX dataset. We also include the performance when the video modality is not included (i.e., ASR task) and we see that the performance highly deteriorates as the noise increases. Instead, the additional use of the video tokens is beneficial as it compensates for the noisy audio tokens. The best configurations suggest that \texttt{Llama-AVSR} can tolerate higher compression rates for the audio tokens (up to $4$) compared to video tokens, both for the clean and noisy settings. This is supported by the results presented in Figure \ref{fig:downsampling}, which revealed that audio tokens can tolerate a higher $K$ compared to video, and by the fact that in extreme noisy conditions (SNR level = $-2$/$5$ dB) the LLM relies more on video tokens. Finally, we find that without compressing the video tokens (i.e., the last row), we obtain no performance gain except in the case of severe noise conditions. 

\begin{table}[t]
\renewcommand{\arraystretch}{1.2}
\renewcommand{\tabcolsep}{2.5mm}
\centering
    \caption{Results of \texttt{Llama-AVSR} for various audio and video compression rates in noisy conditions. $\infty$ means no noise is injected.}

\resizebox{0.85\linewidth}{!}{
\begin{tabular}{cccccccc}

\toprule
\multicolumn{2}{c}{\cellcolor{decorrcolor}\textbf{Compression Rate}} 
 &\multicolumn{6}{c}{\cellcolor{contrcolor}\textbf{SNR Level (dB)}} \\
  \cmidrule(rl){1-2} \cmidrule(rl){3-8}
\bf{A} & \bf{V} & $\infty$  & 5 & 2 & 0 & -2 & -5\\

\midrule
\rowcolor{pastelred}
3 & / & 1.1 & 3.0 & 6.5 & 12.3 & 23.4 & 63.1 \\
\hdashline \addlinespace[2pt]
2 & 4 & 1.0 & 2.3 & 3.9& 4.5 & 10.0 & 18.2   \\
3 & 3 & 1.1 & 2.2& 3.9 & 4.4 & 10.0 & 17.6  \\
 \rowcolor{teagreen}
4 & 2 & \textbf{0.9} & 2.2& 3.8 &\textbf{4.2}  & \textbf{9.5} & 16.9\\
\rowcolor{teagreen}
4 & 1 & \textbf{0.9} & 2.3 & \textbf{3.7} & \textbf{4.2} & \textbf{9.5} & \textbf{16.4} \\

\bottomrule
 \end{tabular}}
\label{tab:ablation_AVSR}
\vspace{-0.5cm}
\end{table}

\section{Conclusion}
\label{sec:conclusion}
In this paper, we present \texttt{Llama-AVSR}, a Multimodal LLM that leverages pre-trained audio/video encoders and an LLM to carry out the tasks of ASR, VSR, and AVSR. By training only lightweight projectors and LoRA modules, we are able to endow a pre-trained LLM with audio and visual speech recognition abilities. 
\texttt{Llama-AVSR} achieves new state-of-the-art results on the LRS3 dataset for the tasks of ASR and AVSR, and comparable performance with previous works for VSR. We also shed light on the key factors that contribute to the powerful results obtained by \texttt{Llama-AVSR}: the choice of the pre-trained encoders and LLM, the use of LoRA modules, and the optimal audio and video compression rates. 

\clearpage

\bibliographystyle{IEEEtran}
\bibliography{refs}

\end{document}